\documentclass[conference]{IEEEtran}
\IEEEoverridecommandlockouts
\usepackage{cite}
\usepackage{amsmath,amssymb,amsfonts}
\usepackage{algorithmic}
\usepackage{graphicx}
\usepackage{textcomp}
\usepackage{xcolor}
\def\BibTeX{{\rm B\kern-.05em{\sc i\kern-.025em b}\kern-.08em
    T\kern-.1667em\lower.7ex\hbox{E}\kern-.125emX}}
\begin{document}

\title{Evaluating Quality of Gaming Narratives Co-created with AI \\
}


\author{\IEEEauthorblockN{Arturo Valdivia}
\IEEEauthorblockA{\textit{Play, Culture and AI Section} \\
\textit{IT University of Copenhagen}\\
Copenhagen, Denmark \\
arturo@valdivia.xyz}
\and
\IEEEauthorblockN{Paolo Burelli}
\IEEEauthorblockA{\textit{Play, Culture and AI Section} \\
IT University of Copenhagen\\
Copenhagen, Denmark \\
pabu@itu.dk}
}

\maketitle

\begin{abstract}
This paper proposes a structured methodology to evaluate AI-generated game narratives, leveraging the Delphi study structure with a panel of narrative design experts. Our approach synthesizes story quality dimensions from literature and expert insights, mapping them into the Kano model framework to understand their impact on player satisfaction. The results can inform game developers on prioritizing quality aspects when co-creating game narratives with generative AI.
\end{abstract}


\section{Introduction}
While \textit{generative AI} has surged into public and research consciousness following the release of systems like ChatGPT, video games have a longer tradition of using AI techniques to generate content that would otherwise be authored by human designers. This tradition is well established in the field of \textit{Procedural Content Generation}, which encompasses a range of methods for algorithmically creating game elements such as levels, characters, quests, and storylines~\cite{nelson2016pcg}. However, the advent of Large Language Models (LLMs) has introduced new levels of sophistication and adaptability to the co-creation of story content, enabling systems that can respond more flexibly to player input, adapt to context, and simulate human-like narrative reasoning.

Despite its promise, it could be argued that the development of AI-assisted narrative generation has outpaced game developers' ability to ensure consistent quality across all LLM-generated outputs. In some cases, poorly generated dialogue or incoherent narrative arcs may disrupt player immersion or actively harm player experience---undermining the very engagement such systems are designed to foster. This \textit{quality gap} becomes especially critical in story-driven games, where narrative is the primary differentiator, and in experiences targeted at sensitive audiences such as children (\textit{e.g.}, games like \textit{YOLI}~\cite{yoli}). Even with well-crafted prompts, the quality of the generated content can vary unpredictably, leading to results that are unconvincing, repetitive, or misaligned with the desired tone or character development. Consequently, ensuring the acceptability of LLM-generated narrative output has become a high-stakes concern for game developers and narrative designers.

In this context, the present work aims to propose a framework to systematically evaluate the quality of gaming narratives co-created with AI. This framework can help game developers to identify what story quality dimensions are most relevant for LLM-generated narratives in their games, and anticipating how do these dimensions map to player satisfaction.

\section{Usage of LLMs for story content in games}

A growing body of research has explored the use of LLMs to generate in-game dialogue, particularly for \textit{Non-player Characters} (NPCs), focusing on aspects such as context-awareness~\cite{csepregi2021contextaware}, narrative coherence~\cite{akoury2023framework}, and character personality~\cite{latouche2023scripts}, with efforts also targeting stylized or story-focused dialogue~\cite{muller2023chatter}. These systems have been used both within games and in external community settings, such as Discord, where NPCs interact with players in ongoing storylines~\cite{sun2023fictional}. Results from these studies show that LLM-generated dialogue can enhance player experience, especially when the models are contextually grounded or guided by structured design frameworks~\cite{ashby2023quest}; although, human-authored dialogue often remains preferred in some contexts~\cite{akoury2023dialogue}.

Beyond dialogue, LLMs are also being employed to co-create broader narrative structures, including story arcs and quests. Research has examined the use of LLMs in generating interactive stories~\cite{sun2023language}, procedural quests~\cite{ashby2023quest}, and complete story games~\cite{yong2023story}, with some projects even characterizing such games as “AI-native”, \textit{i.e.}, games in which generative AI fundamentally shapes gameplay~\cite{sun2023language}. These applications offer both promise and challenge. Although players often appreciate the increased flexibility and personalization enabled by AI-generated content, concerns remain about coherence, consistency, and unintended bias in story outcomes~\cite{gursesli2023climate}, with some studies identifying trends toward overly positive or uninspired endings~\cite{taveekitworachai2023bias}, or uneven output quality across sessions~\cite{taveekitworachai2023journey}.

One pressing question for game teams is: \textit{How can we reliably evaluate the quality of AI-generated narrative content, especially in real-time, where a human narrative designer cannot intervene?}
A natural candidate is the technique known as \textit{LLM-as-a-Judge}, where a second LLM is prompted to evaluate outputs generated by another model. This approach has gained traction due to its scalability and promising correlation with human judgments in creative tasks like storytelling, provided an appropriate prompt design~\cite{yang2024storyeval, gu2025surveyllmasajudge}. However, it raises a second, equally important question: \textit{What should such a judge evaluate?} 
In practice, unguided evaluations often produce unreliable assessments or overly generic feedback.

\section{Proposed Framework to Assess Story Quality}

To ensure narrative quality in LLM-generated stories, we propose a structured evaluation framework involving three stages. First, compiling a list of variables that may have an impact on the quality of the story ---we refer to these variables as \textit{story quality dimensions}. Second, validating the relevance and exhaustiveness of this list through a \textit{Delphi study}~\cite{Ziglio} leveraging a panel of experts on narrative design for games and other more technical game practitioners actively engaged in the generation of stories with LLMs. Third, classifying the quality dimensions in one of the different requirement types (\textit{e.g.}, "\textit{Must-have}" or "\textit{Indifferent}") established in the Kano map~\cite{KANO1984KJ00002952366}. We develop each of these stages in the following sections.


\begin{figure}[htbp]
\centerline{\includegraphics[width=1.02\linewidth]{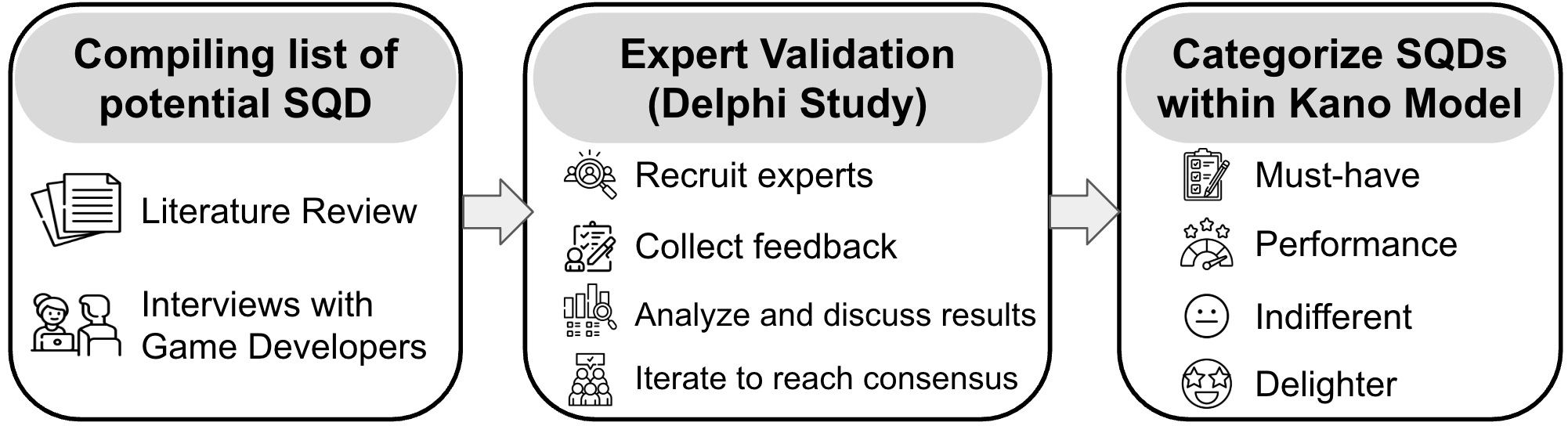}}
\caption{Illustration of the proposed framework to evaluate story quality.}
\label{framework}
\end{figure}

\subsection{Initial list of story quality dimensions}

\begin{figure}[htbp]
\centerline{\includegraphics[width=1.00\linewidth]{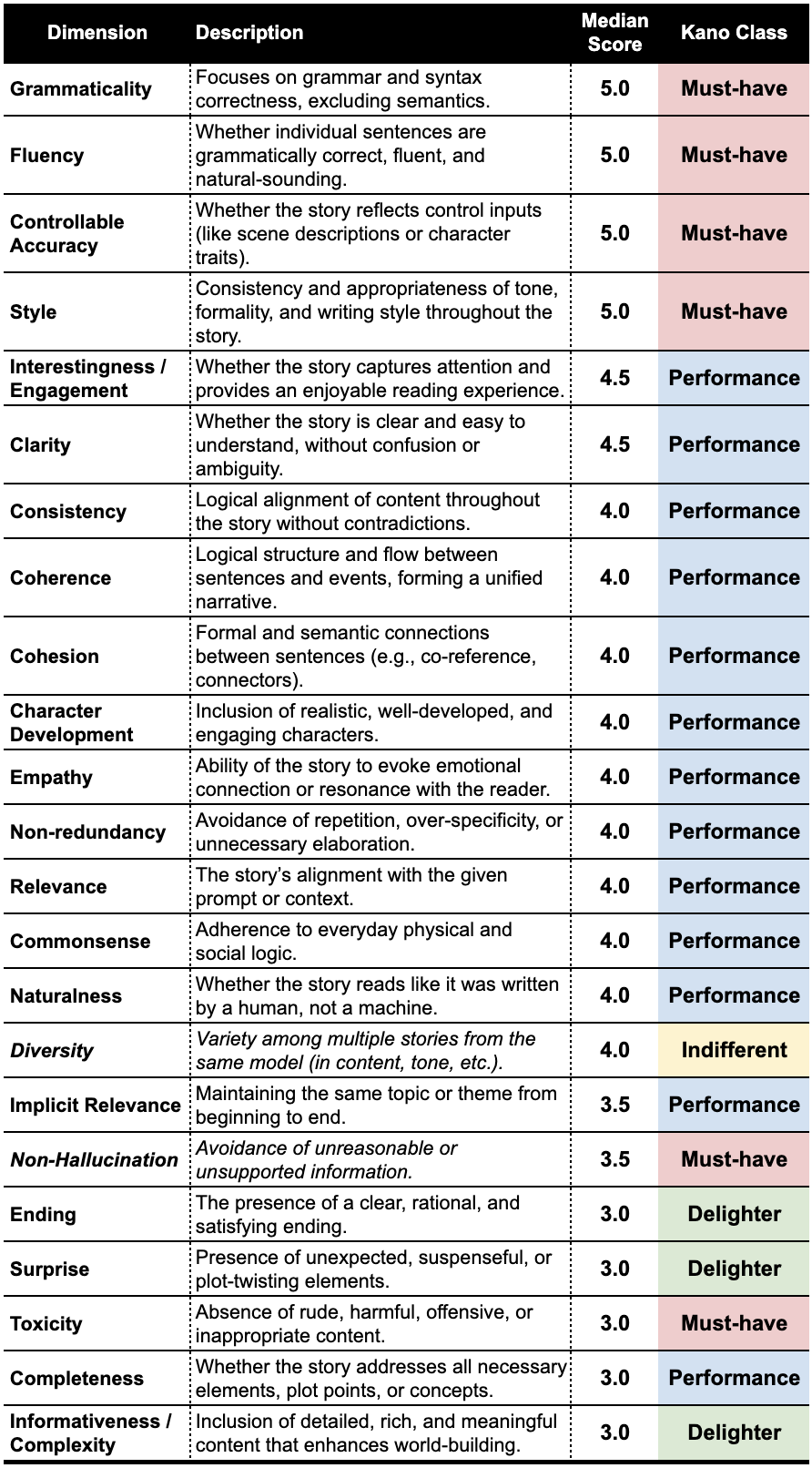}}
\caption{Results from the first round of a Delphi study. The score reflects the median score obtained by surveying our panel of experts. The importance of each Story Quality Dimension is given in a Likert scale from 1 to 5, where 1 means "Not important at all" and 5 means "Critically important". The last column shows the predicted Kano category of player satisfaction.}
\label{table_1}
\end{figure}

As a starting point for compiling a comprehensive list of story quality dimensions (SQDs) ---\textit{i.e.}, variables that may have an impact on the quality of the story--- we utilize the systematic literature review conducted by~\cite{yang2024storyeval}, which identified a list of twenty-three common dimensions used for evaluating story quality. Figure \ref{table_1} contains the list and description of these twenty-three SQDs.

The SQDs outlined in Figure \ref{table_1} serve as overarching categories that encompass various aspects of story quality. Moreover, by drilling down to a second-order level, we can identify the specific factors ---or, \textit{sub-dimensions}--- that contribute to each SQD's overall impact on the story. For instance, in the case of the \textit{Surprise} dimension, the authors in~\cite{Chieppe_Surprise} identify 7 sub-dimensions that make a well-made surprise. These sub-dimensions include for example \textit{Divergence} (the 
magnitude of the knowledge revision caused by the reveal) and \textit{Suddenness} (the speed at which the audience's understanding shifts in response to a new revelation). In this work, however, we focus the analysis exclusively on the first-order SQDs.

It is also worth noting at this point that for some SQDs, there are known quantitative evaluation methods. For instance, in the case of the  \textit{Grammaticality} dimension, the authors in~\cite{Heilman_Grammaticality} employ $\ell_2$-regularized linear regression to learn a model of sentence grammaticality from various linguistic features. Their results suggest a strong agreement with the assessment made by human expert annotators. The evaluation ---whether quantitative or qualitative--- of the different SQDs is considered outside the scope of this work.

\subsection{Expert Validation via Delphi Method}

To validate and refine the list of story quality dimensions, we employ a \textit{Delphi study}. In this section, we primarily follow the framework outlined by~\cite{Ziglio} to introduce the methodology. A Delphi study is a structured, iterative process designed to gather and distill knowledge from a panel of experts through rounds of questionnaires and controlled feedback. This approach aims to generate reliable and creative insights or to produce sustainable information for decision-making, making it especially well-suited to exploratory studies in domains with limited empirical data.

Delphi studies are inherently iterative and typically proceed through multiple rounds to achieve consensus. In each round, participants anonymously provide their opinions via questionnaires. Researchers then collect, synthesize, and return feedback that includes aggregated group responses alongside individualized reflections. This controlled feedback mechanism enables participants to refine their perspectives in light of the broader panel’s input, fostering convergence toward a shared understanding or consensus. Iterations are typically concluded upon observing minimal changes in responses between successive rounds. Anonymity is a crucial element of this process, minimizing the influence of dominant individuals, reducing biases and social pressures, and encouraging candid self-expression and self-correction.

The selection of the expert panel is another critical component. Experts are typically geographically dispersed and chosen for their deep knowledge and practical experience in the relevant domain. A diverse panel ensures a breadth of perspectives, while careful documentation of expert profiles allows readers to assess the credibility and representativeness of the findings. For this study, we engaged a panel of $n=10$ experts in narrative design for games and other technical game practitioners actively working with LLM-generated stories. These experts come from a range of contexts, including AAA studios, top-grossing free-to-play titles, and indie developers. Our panel is distributed across Europe and the United States and has a balanced gender representation. We highlight here that our panel size of experts is within the methodological recommendations made by~\cite{linstone2002delphi}.

Among the various forms of Delphi studies, the \textit{ranking-type Delphi} is particularly well-suited to our work. This variant is structured to reach group consensus on the relative importance of a set of issues and typically proceeds through three stages: brainstorming to identify relevant factors, narrowing down the initial list, and ranking the refined items to identify priorities and consensus. In our study, rather than simply assigning ordinal ranks to the SQDs, experts are asked to categorize each item within one of the requirement types of the \textit{Kano map}, as described in the next section.


\subsection{Player Satisfaction and the Kano Mapping}

To understand how the identified dimensions of the quality of the story influence the reader's satisfaction, we employ principles from the \textit{Kano model}~\cite{KANO1984KJ00002952366}. Originally developed for customer requirement analysis, the Kano model provides a structured framework for examining how the presence or absence of specific attributes affects \textit{customer satisfaction} ---or, in our case, \textit{player satisfaction}. This makes it particularly relevant for our work, as it allows us to move beyond identifying dimensions of story quality to understanding their impact on player experience.

The model uses functional (the feature is present) and dysfunctional (the feature is absent) questions to evaluate the attributes. Based on participants' responses to these paired questions, attributes (here corresponding to story quality dimensions) can be classified into one of the \textit{Kano categories}:

\begin{itemize}
    \item \textbf{Delighter:} Attributes that significantly increase satisfaction when present but do not cause dissatisfaction when absent. These often represent unexpected positive features.
    \item \textbf{Performance:} Attributes whose presence leads to sa\-tisfaction and whose absence results in dissatisfaction. Satisfaction is proportional to the performance of these attributes.
    \item \textbf{Must-have:} Attributes that are considered basic expectations. Their absence or poor performance results in significant dissatisfaction, but their presence does not notably increase satisfaction.
    \item \textbf{Indifferent:} Attributes that neither increase nor decrease satisfaction. Participants are neutral towards these factors.
    \item \textbf{Reverse:} Attributes that cause dissatisfaction when present and satisfaction when absent. These indicate features that are actively disliked.
\end{itemize}


In our study, applying the Kano model framework allows us to link the quality dimensions of the story with their potential effects on the satisfaction of the player. Rather than simply ranking these dimensions, our approach involves asking experts to classify each dimension within one of the Kano categories. This classification provides a nuanced understanding of how different aspects of story quality contribute to overall player experience, offering a more structured and quantifiable perspective on the relative importance of each dimension.

\section{The Results}
Figure \ref{table_1} summarizes the results of our study. The experts participating in this study were asked to rank the importance of each Story Quality Dimension using a Likert scale from 1 to 5, where 1 means ``\textit{Not important at all}'' and 5 means ``\textit{Critically important}''.

In this first round of the Delphi study, none of the twenty-three Story Quality Dimensions (SQDs) received a low median score below 3.0, indicating that no dimension was considered ``\textit{Not important at all}'' or ``\textit{Slightly important}''. Furthermore, 78\% of the SQDs had a median importance score of at least 3.5, placing most SQDs at least as ``\textit{Very important}'', with 26\% of them receiving a median importance score above 4.5, corresponding to ``\textit{Critically important}''. 

On the other hand, we observe that more than half (57\%) of SQDs are expected to be in the ``\textit{One-dimensional}'' category of the Kano mapping, meaning that player satisfaction is expected to be proportional to the level at which the game story is able to perform in each of these dimensions. Additionally, 26\% SQDs were classified as ``\textit{Most-haves}'', 13\% as ``\textit{Attractive}'', and only one SQD was classified as ``\textit{Indifferent}''.

Altogether, these results provide evidence in favor of the relevance of the initial list of SQDs considered for this study.

\section{Discussion and Future Work}
From this first round of the Delphi study, we conclude that the initial list of twenty-three Story Quality Dimension is relevant, and it can be used to inform game developer's prioritization when it comes to decide which aspects of quality should be ensured in the stories generated partially or fully using AI.
An added value from this first round is that feedback gathered from our expert panel provided valuable insights with the potential to extend SQDs beyond the initial list identified by~\cite{yang2024storyeval}. Notably, experts emphasized the importance of two additional dimensions that had not been explicitly considered in the earlier stages of our study: \textit{voice} and \textit{genre alignment}.

\subsection{Emergent dimensions: Voice and Genre Alignment}

First, experts noted that while ``\textit{Naturalness}'' was included as a dimension, it did not fully capture the nuanced aspects of \textit{voice} or \textit{tone}. According to our experts, a distinctive and compelling narrative voice is a critical marker of story quality, distinct from plot or character development. This voice shapes the reader’s engagement and can distinguish a high-quality story from one that is merely functional. As explicitly highlighted by one expert, ``\textit{a good writer should be able to create a unique and compelling voice outside of character and plot elements}''. Based on this input, we recognized \textit{voice} as an independent dimension, classified by the expert as both important and unexpected when present—a marker of particularly high-quality narratives.

Second, experts raised concerns about the degree to which the story aligns with, subverts, or engages with genre conventions. They argued that genre alignment is a fundamental dimension of quality, particularly in contexts where expectations around genre play a critical role in shaping reader satisfaction. For instance, a story that fails to adhere to or meaningfully challenge the tropes of its genre, such as producing a hard-boiled detective story when a fantasy narrative is requested, was identified as a significant quality failure, regardless of technical competence. As such, \textit{genre alignment} emerged as a ``must-be'' requirement for the readers.

By incorporating \textit{voice} and \textit{genre alignment} as distinct dimensions, we expand the evaluative framework to better capture the expectations and nuanced assessments of narrative experts. 

In a forthcoming paper, we conduct a second round of this Dephi study, and we complement it with large scale survey aiming at capturing players' preferences (in the sense of Kano) on a specific game genre. 

\section*{Acknowledgment}
We would like to thank all the experts who participated in this study for the time they generously devoted to completing the surveys and for all the feedback provided during the interviews.

\vfill\null

\bibliographystyle{IEEEtran}

\bibliography{references}

\end{document}